\documentclass[10pt,twocolumn,letterpaper]{article}

\usepackage{iccv}
\usepackage{times}
\usepackage{epsfig}
\usepackage{graphicx}
\usepackage{amsmath}
\usepackage{amssymb}

\usepackage{subfigure}
\usepackage{upgreek}
\usepackage{algorithm}
\usepackage{algorithmic}
\usepackage[multiple]{footmisc}
\usepackage[breaklinks=true,bookmarks=false]{hyperref}

\iccvfinalcopy % *** Uncomment this line for the final submission

\def\iccvPaperID{9482} % *** Enter the ICCV Paper ID here

\ificcvfinal\fi

\begin{document}

%%%%%%%%% TITLE
\title{Search For Deep Graph Neural Networks}

\author{
Guosheng Feng\textsuperscript{1} \quad
Chunnan Wang\textsuperscript{1} \quad 
Hongzhi Wang\textsuperscript{1,2}\\
Harbin Institute of Technology\textsuperscript{1} \quad Peng Cheng Laboratory\textsuperscript{2} \\
{\tt\small 1170300529@stu.hit.edu.cn, WangChunnan@hit.edu.cn, wangzh@hit.edu.cn}
}

\maketitle
\ificcvfinal\thispagestyle{empty}\fi

%%%%%%%%% ABSTRACT
\begin{abstract}
Current GNN oriented NAS methods focus on the search for different layer aggregate components with shallow and simple architectures, which are limited by the ``over-smooth'' problem.
%Although graph neural networks (GNNs) has shown its effectiveness when dealing with non-Euclidean data, robust GNN architectures often require expertise and elaborate design. Meanwhile, Neural Network Architecture Search (NAS) has been successfully applied to design model architectures in computer vision and natural language processing domain. Thus, some efforts has been devoted to discover efficient GNN models using NAS, and achieved promising results. Despite their success, current GNN oriented NAS methods focus on the search for different layer aggregate components with shallow and simple architectures, which are limited by the ``over-smooth'' problem.

To further explore the benefits from structural diversity and depth of GNN architectures, we propose a GNN generation pipeline with a novel two-stage search space, which aims at automatically generating high-performance while transferable deep GNN models in a block-wise manner. Meanwhile, to alleviate the ``over-smooth'' problem, we incorporate multiple flexible residual connection in our search space and apply identity mapping in the basic GNN layers. For the search algorithm, we use deep-q-learning with epsilon-greedy exploration strategy and reward reshaping. Extensive experiments on real-world datasets show that our generated GNN models outperforms existing manually designed and NAS-based ones.
\end{abstract}

%%%%%%%%% BODY TEXT
%---------------------------------------------------------------------Introduction---------------------------------------------------------------------
\section{Introduction}

Graph Neural Networks (GNNs), a class of deep learning models designed for performing information interaction on non-Euclidean graph data, have been successfully applied to applications of various domains such as recommender systems \cite{DBLP:conf/kdd/YingHCEHL18, DBLP:conf/nips/MontiBB17}, physics \cite{DBLP:conf/nips/BattagliaPLRK16, DBLP:conf/icml/Sanchez-Gonzalez18,DBLP:conf/iclr/SeoML20} and natural language processing \cite{DBLP:journals/tacl/ZayatsO18, DBLP:conf/iclr/Johnson17,DBLP:conf/emnlp/LiuLH18}. Although GNNs and their variants \cite{DBLP:conf/nips/HamiltonYL17,DBLP:conf/iclr/KipfW17,DBLP:conf/iclr/VelickovicCCRLB18,DBLP:conf/iclr/XuHLJ19} have shown convincing performance in a variety of graph datasets, the design of graph convolutional layers often require domain knowledge and laborious human involvement. Moreover, unlike most CNN models in computer vision tasks, GNNs often show poor transferability toward different scenarios, and require additional manual adjustments when dealing with new tasks.

Meanwhile, emerged as a powerful tool for automatic neural network optimization, neural network architecture search (NAS) \cite{DBLP:conf/iclr/ZophL17,DBLP:conf/icml/PhamGZLD18,DBLP:conf/nips/NaymanNRFJZ19,DBLP:conf/iclr/LiuSY19,DBLP:conf/eccv/LiuZNSHLFYHM18} has been widely applied for discovering high-performance network architectures in convolutional neural networks (CNNs) and recurrent neural networks (RNNs). Recently, GraphNAS \cite{DBLP:conf/ijcai/GaoYZ0H20} has made the first attempt to apply NAS method to graph data by utilizing existing GNN layers to build the final architecture while leveraging the power of RNN with policy gradient to adaptively optimize the model, followed by several related works as Auto-GNN \cite{Zhou2019AutoGNNNA} and SNAG \cite{DBLP:conf/cikm/ZhaoWY20}.

However, both the hand-crafted and NAS-designed GNN models are constrained by ``over-smooth'' problem \cite{DBLP:conf/aaai/ChenLLLZS20} thus share a simple and shallow architecture. For example, GCN \cite{DBLP:conf/iclr/KipfW17}, GAT \cite{DBLP:conf/iclr/VelickovicCCRLB18} and GraphNAS \cite{DBLP:conf/ijcai/GaoYZ0H20} only stack 2 layers for the transductive learning tasks, while SNAG chooses 3 layers. Although achieving their best performance, such a shallow structure limits the model’s ability to extract information from high-order neighbors, and the potential power brought by structure diversity is underutilized. Some recent works have made efforts to explore different GNN structures, such as Inception-like module \cite{DBLP:journals/corr/abs-2004-11198}, DenseNet-like snowball structure \cite{DBLP:conf/nips/LuanZCP19}, as well as deeper GNN models \cite{DBLP:conf/icml/ChenWHDL20,DBLP:conf/iccv/Li0TG19}. Although these works show possible benefits brought by GNN structural diversity and depth, the structurally diversified GNN design space is not fully explored.

In this work, we aim at digging deeper into the potential benefits of GNN architectures' diversity and depth, and propose a novel GNN generation pipeline to automatically search for high-performance GNN models. As is mentioned in \cite{DBLP:conf/ijcai/GaoYZ0H20}, GNN architecture search suffers from an exponentially growing search space, especially when the layer number increases. To cope with such problem, we propose a two-stage search space that incorporates various flexible model structures such as inception-like module and residual connection, while also greatly reduces the number of candidate models when the overall depth grows. Besides, to further alleviate the ``over-smooth'' problem in deep GNN layers, we apply initial residual connection \cite{DBLP:conf/iclr/KlicperaBG19} and identity mapping \cite{DBLP:conf/icml/ChenWHDL20} to our basic graph convolutional layers. Finally, our search space still contains billions of possible GNN architectures even after pruning. To effectively explore our proposed search space, we apply the deep-q-learning algorithm (DQN) \cite{DBLP:journals/nature/MnihKSRVBGRFOPB15} with epsilon-greedy strategy, accompanied with reward reshaping technique \cite{DBLP:conf/icml/NgHR99} and experience replay \cite{10.5555/168871} that further accelerate and stabilize the search process.

Highlights of our work are listed as follows:
\begin{enumerate}
  \item \textbf{Deep and flexible GNN architecture:} We extend the initial residual connection and identity mapping to various GNN layers thus achieve deep and high-performance GNN models. To fully exploit structural diversity, we also incorporate flexible GNN stacking rules in our search space, which could emulate efficient structures such as residual and inception-like modules.
  \item \textbf{Two-stage search space:} We propose a block-wise GNN architecture generation pipeline, and divide the overall search space into two stages, as block-wise and architecture-wise search, which greatly reduces the exponentially growing search space and improves search efficiency.
  \item \textbf{Transfererble GNN blocks:} Unlike most current GNN structure, our generated GNN block in the first-stage space could be easily transferred to other datasets without researching, while retaining high accuracy.
\end{enumerate}

%---------------------------------------------------------------------Related Work---------------------------------------------------------------------
\section{Related Work}
\subsection{Neural Network Architecture Search}
Reinforcement Learning based neural network architecture search (NAS) was initially proposed by \cite{DBLP:conf/iclr/ZophL17} and \cite{DBLP:conf/iclr/BakerGNR17}, and yield better performance compared with traditional evolutionary based NAS methods \cite{273950,stanley:ec02,6792316,DBLP:conf/ijcai/SuganumaSN18}. However, directly searching for the overall architecture of CNNs and RNNs is computationally expensive, especially in big dataset, as the search space grows exponentially when the scale of architecture increases. Based on this, \cite{DBLP:conf/cvpr/ZhongYWSL18} proposed to break the CNN architectures into blocks, and search the structure of a single block only. The generated block was then used to construct the final architecture. In this work, we propose to apply block-wise search in our first-stage search space, which consists multiple different GNN layers connected with each other in a residual or inception-like manner. Then we use the second-stage search space to control the overall model depth and macro residual connections.
\subsection{Graph Neural Network Architecture}
Designed for learning in graph domains, Graph Neural Networks (GNNs) are initially proposed in \cite{1555942}, with promising performance presented in recent GNN variants \cite{DBLP:conf/iclr/VelickovicCCRLB18,DBLP:conf/iclr/KipfW17,DBLP:conf/nips/HamiltonYL17}. Generally, GNNs could learn node embeddings by performing graph convolutions among adjacent notes, which can be formulated as:
\begin{equation}
	x_i^{(l+1)} = \delta(x_i^{(l)},\ \sigma_{j\in \theta(\mathcal{N}(i))} \gamma(x_i^{(l)}, x_j^{(l)}))
\end{equation}

 Then, inspired by the success of applying NAS technique in CNN and RNN model optimization, several works have attempted to automatically search for good GNN architectures using existing graph convolutional layers \cite{Zhou2019AutoGNNNA, DBLP:conf/ijcai/GaoYZ0H20, DBLP:conf/cikm/ZhaoWY20}. However, these NAS generated GNN models share simple and shallow structures, which underutilizes the information from high-order neighbors. In this work, we propose to search for deep while flexible GNN architecture by incorporating residual and inception-like module in our search space. To cope with the growing search space problem in \cite{DBLP:conf/ijcai/GaoYZ0H20}, we use transferable GNN blocks to construct the final architecture. We also apply initial residual connection \cite{DBLP:conf/iclr/KlicperaBG19} and identity mapping \cite{DBLP:conf/icml/ChenWHDL20} to further alleviate the ``over-smooth'' problem. In experiments, our model consisting of 30 layers reaches the best performance, and both our block and final architecture show great transferability toward different tasks and datasets.

% More recently, \cite{Li2020AutoGraphAG} proposed to search for deep GNN architectures, and utilize multiple skip-connection between GNN layers to alleviate the ``over-smooth'' problem. Despite achieving reasonable performance, we believe the ``over-smooth'' problem could not be properly addressed by simply allowing flexible residual connection. In fact, the best GNN model was achieved at the layer number of 4, and decreases when further stacking to 9 layers, which makes the performance of deep models in \cite{Li2020AutoGraphAG} similar to the NAS generated shallow GNNs as in \cite{DBLP:conf/ijcai/GaoYZ0H20}.

%---------------------------------------------------------------------Methodology---------------------------------------------------------------------
\section{Methodology}
In this section, we first define our problem and search objective. Then we present a general representation of GNN layers with identity mapping and initial residual connection, in order to extend these two technique to various graph convolutional layers. After this, we introduce our proposed two-stage search space with corresponding basic components. Finally, we elaborate our deep-q-learning agent equipped with experience replay and reward reshaping, and illustrate the search process in each stage.
%---------------------------------Problem Formulation---------------------------------
\subsection{Problem Formulation}
The overall search object of this work could be formulated as:
\begin{equation}  
M^* = argmax_{M \in{S}}\ \mathbb{E}[R_V(M)]
\end{equation}
where $S$ is the overall two-stage search space consists of $S_B$ and $S_A$ denoting block-wise and architecture-wise search space respectively. $R_V(M)$ denotes the model's evaluated accuracy (reward) on the validation set $V$. The aim of this work is to find the best GNN model $M^* \in S$ that achieves highest expected accuracy $\mathbb{E}[R_V(M^*)]$.

Meanwhile, as our model generation is a progressive process among the two-stage search space, we can not directly excavate the best model $M^*$ in $S$. Thus, we further define the sub-goal of architecture search in each stage as:
\begin{align}
B^* = argmax_{B \in{S_B}}\ \mathbb{E}[R_V(M'_B)] \\
A^* = argmax_{A \in{S_A}}\ \mathbb{E}[R_V(M_A)]
\end{align}
where $B$ and $A$ denote for GNN block and architecture respectively, while each $B$ is composed by basic GNN layers and $A$ could contain multiple kinds of generated high-performance GNN blocks. Note that $M'_B$ is the standard model constructed by block $B$, as depicted in Figure \ref{StandardArchitecture} , with similar architecture depth defined in the second search space. And $M_A$ is the GNN model directly decided by architecture $A$.

%---------------------------------Two-stage Search Space---------------------------------
\subsection{Identity Mapping \& Initial Residual Connect} \label{section_IdentityInitial}
While searching for deep GNN architecture, our model could severely suffer from the ``over-smooth'' problem \cite{zhou2020understanding}. Recently, APPNP \cite{DBLP:conf/iclr/KlicperaBG19} applied initial residual connection in the context of Personalized PageRank to alleviate this problem, but failed to achieve a deep model. Following \cite{DBLP:conf/iclr/KlicperaBG19}, GCNII \cite{DBLP:conf/icml/ChenWHDL20} further pointed out that solely applying initial residual connection is not sufficient. Inspired by Resnet \cite{DBLP:conf/cvpr/HeZRS16}, they utilized identity mapping as supplement for initial residual connection, which successfully deepened GCN model to 64 layers and achieved top performance. Inspired by these works, we propose to extend these two techniques to other graph convolutional layers, and then utilize these layers as basic components of our GNN blocks. Here, we present a general representation of a GNN layer with identity mapping and initial residual connection as:
\begin{equation}
	\begin{aligned}
	h_{i}^{l+1} = \sigma_{j\in \mathcal{N}(i)}[( (1-\alpha)c_{ij}h_{j}+\alpha h^{(0)}_{j})((1-\beta)+\beta \theta_{ij})]	
	\end{aligned}
\end{equation}
where $h_i^{(l)}$ is the embedding for node $i$ in layer $l$. $\alpha$ and $\beta$ are two hyper-parameters that control the weight for residual connection from the initial embedding $h_j^{(0)}$ and identity mapping for the weight matrix $\theta_{ij}$ respectively.  And $c_{ij}$ denotes the coefficients between target node $i$ and its one-hop adjacent node $j$. For example, $c_{ij}$ is the attention coefficients $a_{ij}$ in GAT \cite{DBLP:conf/iclr/VelickovicCCRLB18} and $\frac{1}{\sqrt{d_id_j}}$ in GCN \cite{DBLP:conf/iclr/KipfW17}. Finally, a permutation invariant aggregation function $\sigma$ is applied to aggregate  features from adjacent nodes.

%---------------------------------Two-stage Search Space---------------------------------
\begin{figure}
\begin{center}
\subfigure[GNN block]{\includegraphics[width=4.5cm,height=6.5cm]{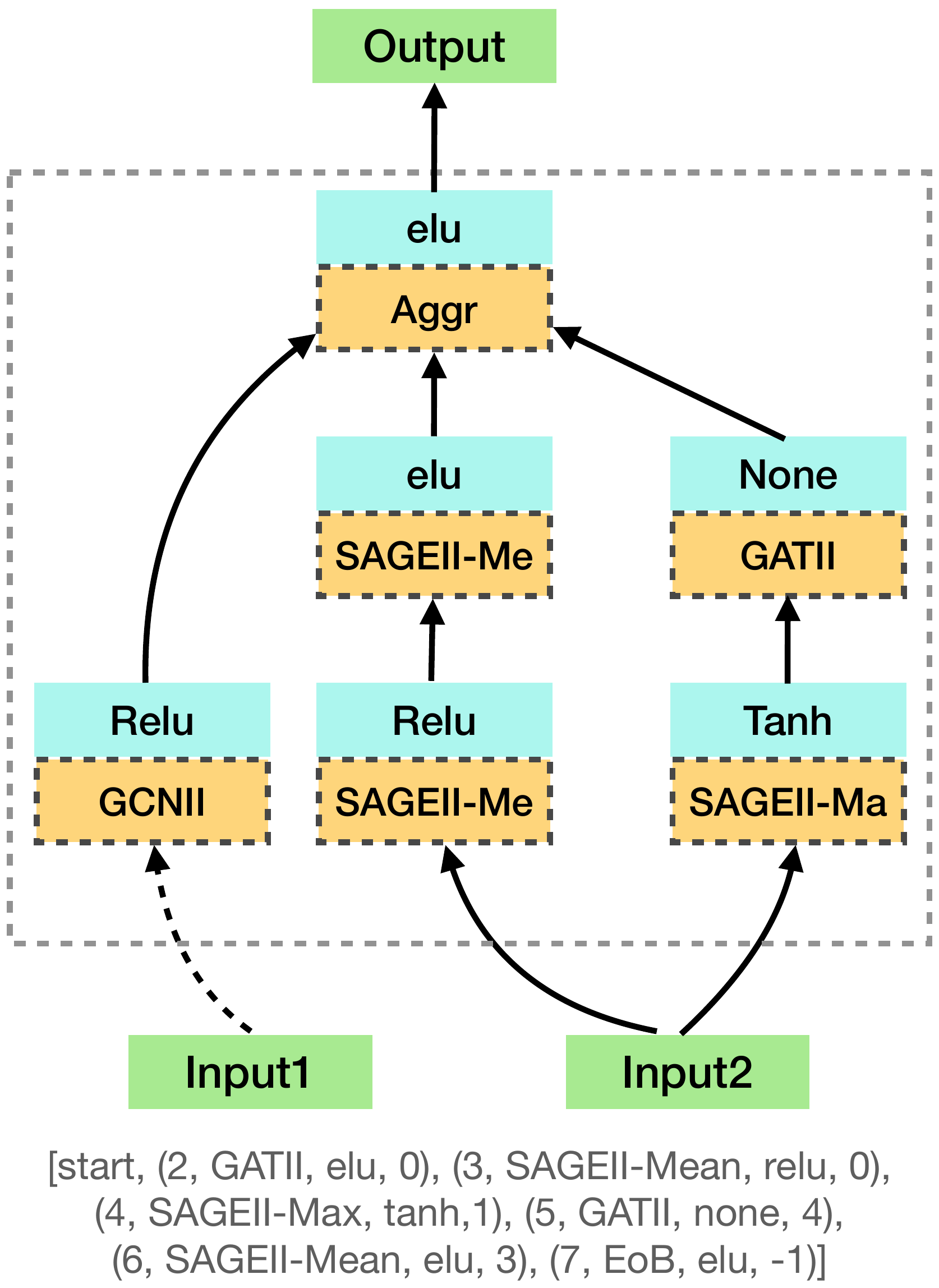}
\label{Block}
}
\subfigure[GNN architecture]{
{\includegraphics[width=3.4cm,height=8cm]{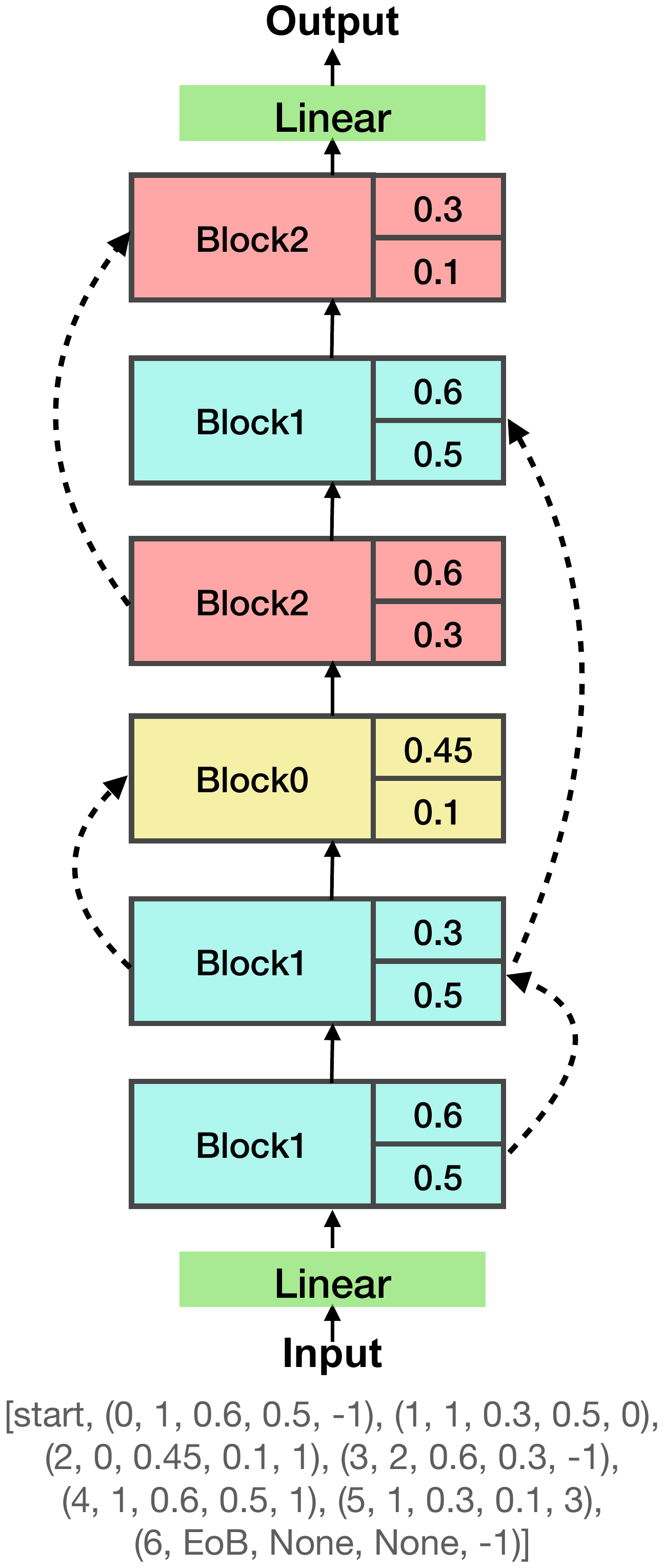}}
\label{Architecture}
}
    \caption{\textit{Left:} Illustration of a block with corresponding representative block code. Each layer in the block is depicted as a 4-D vector. \eg (2, GATII, elu, 0). \textit{Right:}  Schematic  depiction of a GNN architecture constructed by three types of blocks, with corresponding architecture code. Each block is depicted as a 5-D vector. Note that dashed lines with arrows denote for residual connections.}
  
\end{center}
\end{figure}

\subsection{Two-stage Search Space}
In this section, we elaborate our proposed search space for graph neural network generation. Inspired by the design of modern CNNs, \eg Resnet \cite{DBLP:conf/cvpr/HeZRS16} and Inception \cite{DBLP:conf/cvpr/SzegedyLJSRAEVR15}, that stack multiple hand-crafted CNN blocks to construct the final architecture, we divide the search process into two stages as block-wise and architecture-wise search, to scale down the enormous GNN search space. By applying such block-stacking neural network design, our discovered high-performance GNN blocks can be easily transferred to other datasets and tasks without researching, while still achieving state-of-the-art performance.

\textbf{Block-wise search space.} Instead of directly searching for the overall architecture, we start from discovering efficient GNN blocks. To construct a GNN block, several different GNN layers are connected together in a residual or inception-like manner. Similar to CNNs, graph convolutional layers also contain a feed-forward propagation procedure. Thus, we represent the structure of a GNN block using directed acyclic graph (DAG), and depict each layer with corresponding activation function and predecessor by 4-D vectors as:

\[[index, layer type, activation, prefix]\]
which is illustrated in Figure \ref{Block}. Based on such 4-D-vector representation, the overall search options for the block-wise search space are shown in Table \ref{stage1_table}. We use layer \textit{index} to identify each layer, and max layer number $N_l$ to control the size of GNN blocks. To process graph convolutions, we choose five types of commonly used GNN layers equipped with identity mapping and initial residual connection. For convenience, we use the suffix 'II' to denote such modification, \eg GAT with identity mapping and initial residual connection is written as GATII. In practice, we fix the multi-head number to 1, to make GAT compatible with identity mapping. Meanwhile, to allow sizes of blocks variable during search, we add a special 'EoB' layer that denotes the End of a Block. After encountering the 'EoB' layer, all the output of layers without a successor will be aggregated as the final output of the block. Finally, to apply residual connection in blocks, we allow each layer to choose a previous layer (\ie \textit{prefix}) as its predecessor. Note that each block has two inputs, one is the standard input from the previous block, and the other is the macro residual input, which will be discussed in the next sub-section. We add these two inputs as options of layer prefix, noted as -1 and 0 respectively, to make information from shallow layers directly applicable to deep layers.
\begin{table}
\begin{center}
\begin{tabular}{|c|c|c|}
\hline
Layer Components & Options \\
\hline\hline
Index & 1, 2, 3, ... , $N_l$ \\
\hline
GNN Layer & GCNII, GATII, SAGEII-Mean,\\& SAGEII-Max, AGNNII, EoB \\
\hline
Activation Func & ReLU, ELU, PReLU, Tanh, \\& Identity, none\\
\hline
Prefix & -1, 0, 1, ... , \{current index\} -1\\
\hline
\end{tabular}
\end{center}
\caption{Different options of layer components in block-wise search space $S_B$.}
\label{stage1_table}
\end{table}

\begin{table}
\begin{center}
\begin{tabular}{|c|c|c|}
\hline
Block Components & Options \\
\hline\hline
Index & 1, 2, 3, ... , $N_b$ \\
\hline
GNN Block & \{Blocks in GBP\}, EoB \\
\hline
Dropout Rate & 0.3, 0.45, 0.6\\
\hline
Alpha & 0.1, 0.3, 0.5\\
\hline
Prefix & -1, 0, 1, ... , \{current index\} -1\\
\hline
\end{tabular}
\end{center}
\caption{Different options of block components in architecture-wise search space $S_A$.  }
\label{stage2_table}
\end{table}

\textbf{Architecture-wise search space.} After obtaining multiple GNN blocks during the first-stage search, we store the Top-\textit{N} blocks in the \textbf{GNN Block Pool} ($GBP$) that is used to construct the architecture-wise search space, where \textit{N} is manually defined. In this stage, we want the agent to learn efficient GNN block stacking rules thus generate high-performance GNN architecture. Recall that we allow each block to have two inputs. We set one of them as the macro residual input that is chosen from the out of all previous blocks, in order to let information flow from shallow layers to deep layers directly. At the same time, as is considered by \cite{DBLP:conf/icml/ChenWHDL20} as influential parameters when facing different tasks and datasets, the dropout rate and weight for initial residual connections are also added to the architecture-wise search apace,  and thus changeable in different blocks. Above all, we describe each block in the architecture-wise space using 5-D vectors as: 
\[[index, block type, dropout, alpha, prefix]\] 
An example of auto-generated architecture with corresponding architectural code is depicted in Picture \ref{Architecture}, and the overall options for each element in the 5-D vector are shown in Table \ref{stage2_table}. Similar to block-wise search, we use block \textit{index} to identify each block, and use the max block number $N_b$ to control the size of the final architecture. The \textit{block} is chosen from GBP that is obtained during the block-wise search. Meanwhile, we use \textit{prefix} to control the macro residual connections between blocks, where -1 denotes no residual input, and 0 refer to residual input from the first block.

%---------------------------------Q-learning Agent---------------------------------
\begin{figure*}
\centering
\subfigure[Block Sampling Process]{
\includegraphics[width=14cm, height=2.8cm]{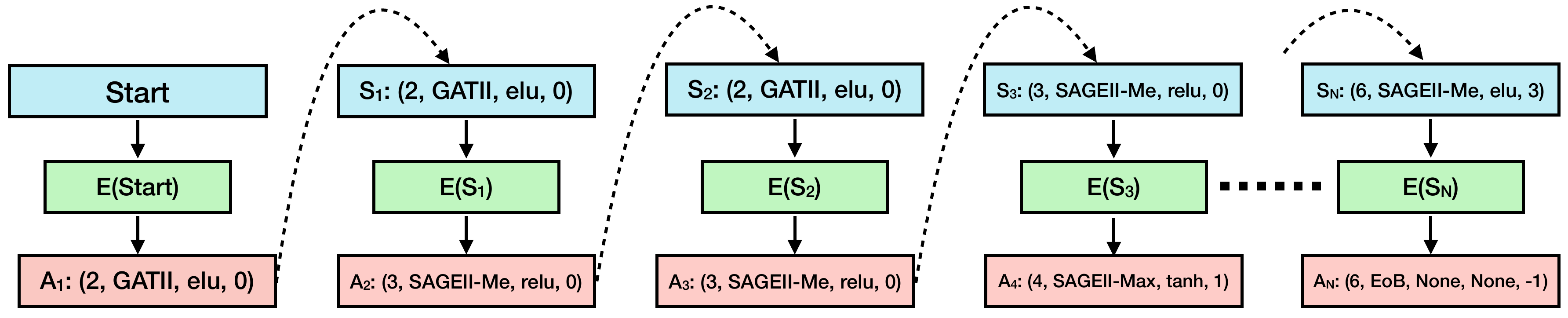}
\label{Sample}
}
\quad
\subfigure[Standard Architecture]{
\includegraphics[width=8.6cm]{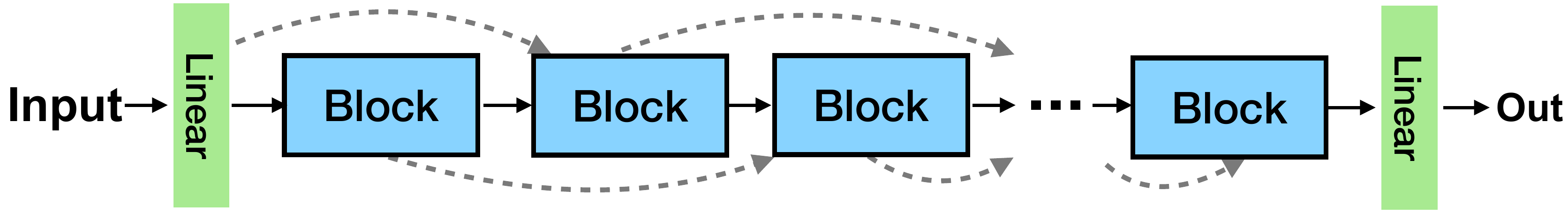}
\label{StandardArchitecture}
}
\quad
\subfigure[Model Generation Pipline]{
\includegraphics[width=7.6cm]{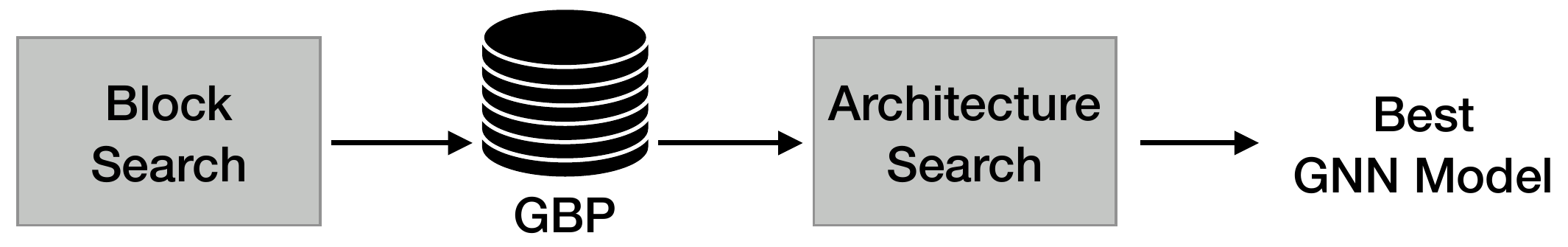}
\label{Pipline}
}
\caption{(a). A example of block sampling process, where $\text{A}_i$ and $\text{S}_j$ stand for the i-th action and state respectively, and E(S) denotes for the set of valid actions in state S. (b). The standard architecture used for GNN block evaluation. Each block have two inputs, the direct input from previous block and the residual input from the block right before the previous block. (c). Illustration of our proposed GNN generation pipline. More details about the search process are shown in Algorithm \ref{algo_search_block}.}
\end{figure*}

\begin{algorithm}
\caption{DQN for Block-wise search}
\label{algo_search_block}
\textbf{Given:} block-wise space $S_B$, DQN agent $DQN_\theta$, \\ block pool $GBP$, memory pool $M$, search epoch $\#_B$, \\ epsilon greedy rate $\epsilon$ \\
\textbf{Output:} Top-k block codes
\begin{algorithmic}[1]
\STATE Initiate $DQN$ parameterized by $\theta$
\STATE state = 'start'
\FOR {$i = 1 \to \#_B$}
\STATE $\mathbb{C} = [\ ]$
\WHILE {'EoB' not encountered}
\STATE rand = random(0,1)
\IF {rand $\le \epsilon$}
\STATE action = $DQN_\theta$(state) \quad //exploitation
\ELSE 
\STATE action = random\_sample(state) \quad //exploration
\ENDIF
\STATE state = action
\STATE $\mathbb{C}$.append(state)
\ENDWHILE
\STATE Build model $M_\mathbb{C}$ based on code $\mathbb{C}$
\STATE $R_V(M_\mathbb{C})$ = Evaluate($M_\mathbb{C}$) \quad //train and test model
\STATE Update memory pool $P$
\STATE Update $\theta$ using $batch$ from $P$ based on Equation (\ref{q-function}) and Equation (\ref{q-function-final})
\STATE \textbf{if} $R_V(M_\mathbb{C}) \textgreater min\_acc(GBP)$ \textbf{then} update GBP
\ENDFOR
\RETURN GBP
\end{algorithmic}
\end{algorithm}

\subsection{Deep-q-learning Agent}
Even after division and pruning, our search space still contains billions of candidate blocks and architectures. For example,  when setting the maximum layer number $N_l$ to 6, there are approximately $2.047 \times 10^{18}$ candidate blocks in the block-wise search space. As GNN model evaluation is time-consuming, it is impossible to iterate over all candidate models. Thus, we need an accurate while sample-efficient search agent to explore the search space. Here, we apply deep-q-learning \cite{DBLP:journals/nature/MnihKSRVBGRFOPB15}, a variant of Q-learning algorithm, to guide the search over our proposed search space. For better illustration, we use the vanilla Q-learning to describe  our agent definitions. 

Q-learning is an \textit{off-policy} reinforcement learning algorithm, which aims at taking the best action $A$ at each step so as to maximize the expected cumulative reward $\mathbb{E}(R)$. There are four key components for q-learning, as  \textit{agent, environment, state} and \textit{action}.  Since we describe our block or architecture using a list of k-D vectors (\eg k=4 in block-wise search), we define such list of vectors as block/architecture code $\mathbb{C}=[V_1,V_2, ..., V_N]$ where $V_i$ denotes for the i-th k-D vectors. To apply Q-learning algorithm to our search process, we consider the sampling of $\mathbb{C}$ as a Markov Decision Process, which means the generation of each k-D vector is probabilistically independent. Based on such presumption, \textit{states} and \textit{actions} in Q-learning algorithm could be regarded as the k-D vectors in our search space, while a special 'start' state is set as the initial state. An example of block-wise sampling process is shown in Figure \ref{Sample}. Then, the objective of our agent should be discovering the trajectory  $\Gamma_{\mathbb{C}}$ with the highest expected reward: 
\begin{equation}
	R^* = \mathop{argmax}\limits_{\mathbb{C}\in S}(\mathbb{E}[R(\Gamma_\mathbb{C})])
\end{equation}
 Because Q-learning is a kind of Temporal-Difference learning method whose agent could be updated based on an existing estimate, the above objective could be achieved by applying the Bellman Equation recursively. Then, consider time step $t$ and an agent policy $\pi$ , the expected reward in state $S_t$ could be written as:
 \begin{equation}
 	 \begin{aligned}
 R_{\pi}(S_t) &= \mathbb{E}_\pi[\sum_{k=0}^\infty \gamma^kr_{t+k+1}|S_t] \\
 &= \mathbb{E}_\pi[r_{t+1} + \gamma R_\pi^*(S_{t+1})|S_t]
 \end{aligned}
 \end{equation}
where $r_t$ is the instant reward obtained when transiting from $S_{t-1}$ to $S_t$, and $\gamma$ denotes the discount factor that measures the importance of future expected reward. In Q-learning, each expected reward $R_\pi$ in state $S_t$ is regarded as a Q-value of \textit{state-action} pair. Therefore, the above equation could be solved by iteratively updating the Q-value as:
 \begin{equation}
 	 \begin{aligned}
 	Q(S_t,A_t) = &(1-\alpha)Q(S_t,A_t) + \\ &\alpha[r_{t+1}+\gamma \max\limits_{A'\in E(S_t)}Q(S_{t+1}, A')]
 	\label{q-function}
 \end{aligned}
 \end{equation}
where $\alpha$ is the learning rate for updating the Q-value, $r_{t+1}$ is the reshaped instant reward which we will discuss later, and $E(S_t)$ is the set of all valid actions in state $S_t$. Additionally, when encountering the last action $A_T$, the future expected reward term in Equation (\ref{q-function}) is dismissed:
 \begin{align}
 	Q(S_{T-1},A_T) = (1-\alpha)Q(S_{T-1},A_T) + \alpha R_T
 \label{q-function-final}
 \end{align}
where $R_T$ is the final reward (\ie model evaluated  accuracy $R_V(M)$). Note that, because the model could only be constructed and evaluated after the the code $\mathbb{C}$ is generated, the instant reward term $r_t$ is hard to decide. Thus, we use reward reshaping \cite{DBLP:conf/icml/NgHR99} to adjust each $r_t$ as:
\begin{align}
	r_t = R_T/\text{len}(\mathbb{C})
\end{align}

Although above definitions make our search agent theoretically feasible, in practice it is unacceptable when applying vanilla q-learning as there are too many states and actions to be stored in q-table. Therefore, we apply two 2-layer MLPs (\ie deep-q-networks) to approximate the q-table, similar to the deep-q-learning \cite{DBLP:journals/nature/MnihKSRVBGRFOPB15}. Meanwhile, as approximating the \textit{action-value} function  often leads to instability \cite{baird1995residual}, we further apply Experience Replay \cite{10.5555/168871,mnih2015human} to stabilize the search process. The details will be discussed in the next section. The pseudo code for our guided block-wise search is shown in Algorithm \ref{algo_search_block}. Because of space limitations, we do not present the pseudo code for architecture-wise search, which is similar to block-wise search if we replace $GBP$ in Algorithm \ref{algo_search_block} with $best\ model$.
\begin{table*}
\begin{center}
\resizebox{0.9\textwidth}{!}{
\begin{tabular}{|l|cccccccccc|}
\hline
Dataset & Cora & CiteSeer & PubMed & CS & Physics & Computer & Photo & Cornell & Texas & Wisconsin \\
\hline\hline
\# Nodes & 2708 & 3327 & 19717 & 18333 & 34493 & 13381 & 7487 & 183 & 183 & 251 \\

\# Edges & 5429 & 4732 & 44338 & 81894 & 247962 & 245778 & 119043 & 295 & 309 & 499 \\

\# Features & 1433 & 3703 & 500 & 6805 & 8415 & 767 & 745 & 1703 & 1703 & 1703\\

\# Classes & 7 & 6 & 3 & 15 & 5 & 10 & 8 & 5 & 5 & 5\\
\hline
\end{tabular}
}
\end{center}
\caption{Dataset descriptions.}
\label{datasets_table}
\end{table*}
%---------------------------------Experiment---------------------------------
\section{Experiment}
In this section, we validate the performance of our proposed method by comparing our auto-generated GNN models against several state-of-the-art graph neural network models on various benchmark datasets. For convenience, our NAS-generated deep GNN model are denoted as \textbf{Deep-GNAS}\footnote{please see our supplementary material for detailed description of our generated models on different datasets}, while all the directly transferred models and results without researching in the two-stage search space are marked with asterisks.
%---------------------------------Datasets---------------------------------
\subsection{Datasets \& Experimental Setup}
\textbf{Datasets.} We apply our model to 10 open graph benchmarks in Table \ref{datasets_table}. Settings and descriptions for each dataset will be illustrated under different learning scenarios.

\textbf{Settings of DQN Agent.} To approximate the vanilla q-table, we use two 2-layer MLPs as deep-q-networks, one (\ie evaluation network) is responsible for mapping the current state to the next action that maximizes future expected reward as in Equation (\ref{q-function}), and the other (\ie target network) is used to train the former MLP with frozen parameters. We update the parameters of target network every 100 search epochs. For the epsilon-greedy sampling, we set the initial exploration rate to 1 and anneal it to zero following cosine decay schedule \cite{DBLP:conf/iclr/LoshchilovH17} during the last 60\% epochs. To replay the agent memory, we set the memory capacity as 300, and use a batch-size of 32 with future expected reward weight $\gamma=1$. As for the optimizer, we apply Adam \cite{DBLP:journals/corr/KingmaB14} with learning rate 0.01. To filter out outstanding GNN models, We set the epoch for block-wise search to 1500 and the size of GNN block pool $GBP$ to 3. After obtaining the top-3 blocks as shown in Figure \ref{GBP}, we explore the architecture-wise space for 1000 epochs to discover the best GNN model.

\textbf{Settings of GNN models.} For the block-wise search, we set initial residual connection weight $\alpha=0.3$, identity mapping parameter $\lambda=0.5$, dropout rate $=0.3$.  the maximum layer number for each block is set to 6 with hidden size 32. Meanwhile, we use addition for aggregating layer features in a block. Each sampled block is evaluated using the 8-block standard architecture depicted in Figure \ref{StandardArchitecture}, which is trained for 400 epochs using Adam optimizer with learning rate 0.01. Following \cite{DBLP:conf/icml/ChenWHDL20}, we set the weight decay for graph convolutional layers to 0.01 and fully connected layers to 4e-5. For the architecture-wise search, the maximum block number is set to 8 while other settings are specified for different datasets. 

\begin{table}
\begin{center}
\resizebox{0.489\textwidth}{!}{
\begin{tabular}{c|c|ccc}
\hline
Type & Method  & Cora &CiteSeer & PubMed\\
\hline 
& GCN & 81.5 & 71.1 & 79.0\\
& GAT & 83.1  & 70.8 & 78.5\\
& ARMA & 83.4 & 72.5 & 78.9\\
Hand- & APPNP & 83.3 & 71.8 & 80.1\\ 
Crafted & HGCN & 79.8 (9) & 70.0 (9) & 78.4 (9) \\
& JKNet(Drop) & 83.3 (4) & 72.6 (16) & 79.2 (32)\\
& Incep(Drop) & 83.5 (64) & 72.7 (4) & 79.5 (4) \\
& GCNII & 85.5 (64) & 73.4 (32) & 80.2 (16) \\
\hline 
 & GraphNAS-R & 83.3 & 73.4 & 79.0 \\
NAS- & GraphNAS & 83.7 & 73.5 & 80.5 \\
Based & AutoGNN & 83.6 & \textbf{73.8} & 79.7 \\
\cline{2-5}
& DeepGNAS & \textbf{85.6} (28) & 73.6* (31) & \textbf{81.0}* (27) \\ 
\hline
\end{tabular}
}
\end{center}
\caption{Summary of semi-supervised classification accuracy (\%) of different methods on three citation networks. Following \cite{DBLP:conf/icml/ChenWHDL20}, the number in parentheses denotes the depth (\ie layer number) of corresponding GNN model. The results without parentheses are obtained with 2-layer shallow models. And the results with asterisk are obtained using transferred blocks.}
\label{semi_table}
\end{table}

\begin{table*}
\begin{center}
\resizebox{0.8\textwidth}{!}{
\huge
\begin{tabular}{c|c|cccccc}
\hline
Type & Method  & Cora & CiteSeer & PubMed & Cornell & Texas & Wisconsin \\
\hline 
& GCN & 85.77 & 73.68 & 88.13 & 52.70 & 52.16 & 45.88\\
& GAT & 86.37  & 74.32 & 87.62 & 54.32 & 58.38 & 49.41\\
& APPNP & 87.87 & 76.53 & 89.40 & 73.51 & 65.41 & 69.02\\
& Geom-GCN-I & 85.19 & 77.99 & 90.05 & 56.76 &  57.58 & 58.24\\
Hand-Crafted& JKNet & 85.25 (16) & 75.85 (8) & 88.94 (64) & 57.30 (4) & 56.49 (32) & 48.82 (8)\\ 
& JKNet(Drop) & 87.46 (16) & 75.96 (8) & 89.45 (64) & 61.08 (4) & 57.30 (32) & 50.59 (8)\\
& Incep(Drop) & 86.86 (8) & 76.83 (8) & 89.18 (4) & 61.62 (16) & 57.84 (8) & 50.20 (8) \\
& GCNII & 88.49 (64) & 77.08 (64) & 89.57 (64) & 74.86 (16) & 69.46 (32) & 74.12 (16)\\
& GCNII$^\dagger$ & 88.01 (64) & 77.13 (64) & 90.30 (64) & 76.49 (16) & 77.84 (32) & 81.57 (16) \\
\hline
& GraphNAS & 88.40 & 77.62 & 88.96 & - & - & - \\
NAS-Based & SNAG & 88.95 (3) & 76.95 (3) & 89.42 (3) & - & - & - \\ 
\cline{2-8}
& DeepGNAS* & \textbf{91.23} (30) & \textbf{84.04} (31) & \textbf{90.74} (28) & \textbf{91.89} (30) & \textbf{94.59} (30) & \textbf{95.65} (30) \\ 
\hline
\end{tabular}
}
\end{center}
\caption{Summary of full-supervised classification accuracy (\%) of different methods on three citation networks and three webpage datasets. Following \cite{DBLP:conf/icml/ChenWHDL20}, number in parentheses denotes depth (\ie layer number) of corresponding GNN model. Results without parentheses are obtained with 2-layer shallow models. Note that GCNII$^\dagger$ refers to the GCNII variant proposed in \cite{DBLP:conf/icml/ChenWHDL20}. *Our results are obtained from reused blocks generated from semi-supervised first-stage search on Cora dataset.}
\label{full_table}
\end{table*}

\subsection{Semi-supervised Learning}
In the semi-supervised node classification tasks, we test our model on three Citation networks as Cora, PubMet and CiteSeer. In these datasets, documents are represented as graph nodes and edges correspond to citation links; each node feature corresponds to the bag-of-words representation of the corresponding document while each node correspond to a class label. For fair comparison, we apply the standard training/validation/testing split following \cite{DBLP:conf/icml/YangCS16}, where 20 nodes per class are used for training, 500 nodes for validation and 1000 nodes for testing. For baselines, we choose 8 hand-crafted GNN models and 2 NAS-generated models. Among those hand-crafted ones, the first 4 models share shallow 2-layer GNN structures: GCN \cite{DBLP:conf/iclr/KipfW17}, GAT \cite{DBLP:conf/iclr/VelickovicCCRLB18}, ARMA \cite{DBLP:journals/corr/abs-1901-01343}, APPNP \cite{DBLP:conf/iclr/KlicperaBG19}. For the rest manually designed baselines, we follow \cite{DBLP:conf/icml/ChenWHDL20} that equip JKNet \cite{DBLP:conf/icml/XuLTSKJ18} and IncepGCN \cite{DBLP:conf/iclr/RongHXH20} with DropEdge technique, and add one recent hierarchical method HGCN \cite{DBLP:conf/ijcai/Hu0WWT19}, which could serve as deep manual baselines. The metrics are reused as reported in \cite{DBLP:conf/icml/ChenWHDL20} and \cite{DBLP:conf/ijcai/GaoYZ0H20}. Note that unlike previous NAS methods that search for GNN model from scratch on each dataset, we conduct the block-wise search on Cora while transfer the generated blocks in $GBP$ to other datasets, which also leads to competitive results.  The results are shown in Table \ref{semi_table}. We observe that DeepGNAS achieved best performance on Cora and PubMed while seconding to AutoGNN by 0.2\% in CiteSeer dataset. Notably, our model's accuracy on Cora surpassed all NAS-generated models by about 2\%.

\begin{table}
\begin{center}
\resizebox{0.495\textwidth}{!}{
\Huge
\begin{tabular}{c|c|cccc}
\hline
Type & Method  & CS & Physics & Computer & Photo \\
\hline 
& GCN & 95.5$\pm$0.3 & 98.3$\pm$0.2 & 88.0$\pm$0.6 & 95.4$\pm$0.3\\
Hand- & GAT & 95.5$\pm$0.3  & 98.1$\pm$0.2 & 89.1$\pm$0.6 & 95.6$\pm$0.3\\
Crafted & ARMA & 95.4$\pm$0.2 & 98.5$\pm$0.1 & 86.1$\pm$1.0 & 94.8$\pm$0.8\\
& APPNP & 95.6$\pm$0.2 & 98.5$\pm$0.1 & 89.8$\pm$0.4 & 95.8$\pm$0.3\\ 
\hline 
NAS- & GraphNAS & 97.1$\pm$0.2 & 98.5$\pm$0.2 & 92.0$\pm$0.4 & 96.5$\pm$0.4 \\
\cline{2-6}
Based& DeepGNAS & \textbf{98.6$\pm$0.3} & \textbf{99.5$\pm$0.3} & \textbf{92.3$\pm$0.5} & \textbf{96.6$\pm$0.2}\\
\hline
\end{tabular}
}
\end{center}
\caption{Results of model transferability test in terms of accuracy (\%).}
\label{trans_table}
\end{table}

\subsection{Full-supervised Learning}
\begin{figure*}
\centering
\subfigure[Generated blocks stored in $GBP$]{
\includegraphics[width=9cm, height=6cm]{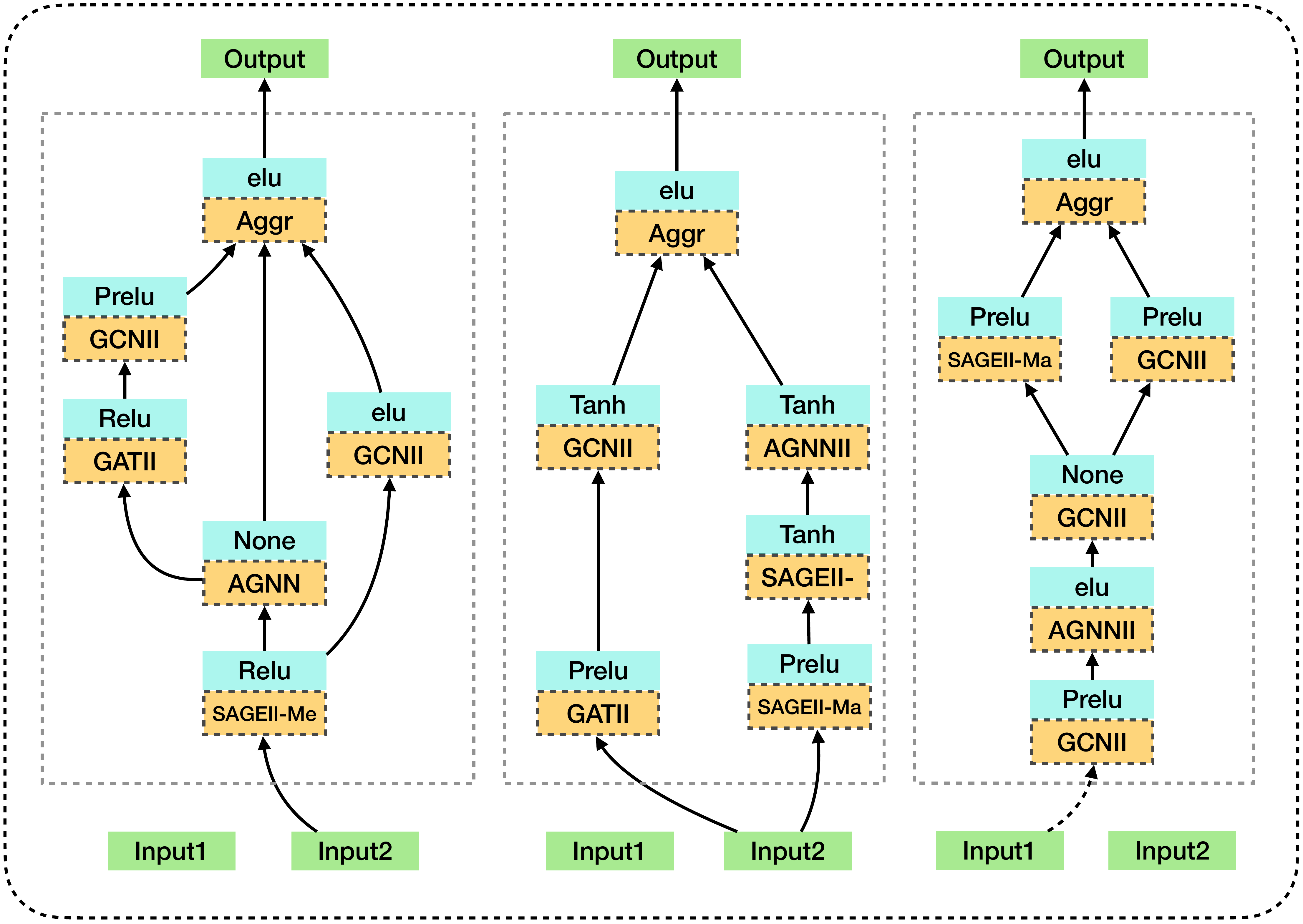}
\label{GBP}
}
\quad
\subfigure[Model-Cora]{
\includegraphics[width=2cm]{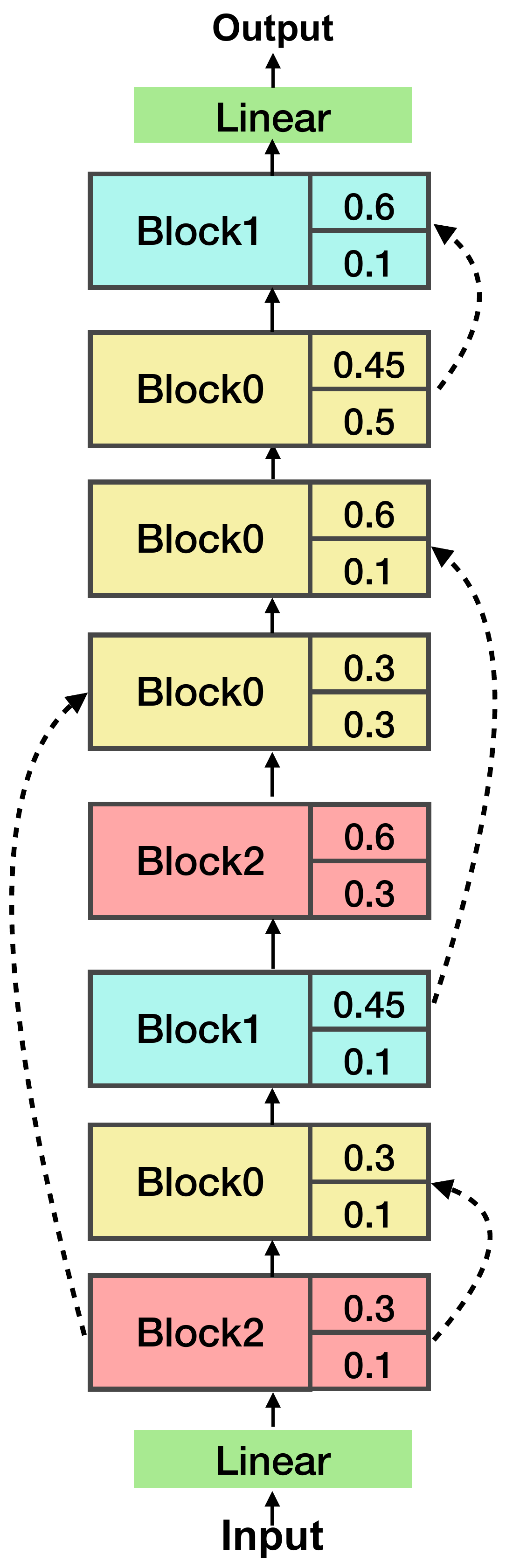}
\label{Model-Cora}
}
\quad
\subfigure[Model-CiteSeer]{
\includegraphics[width=2.23cm]{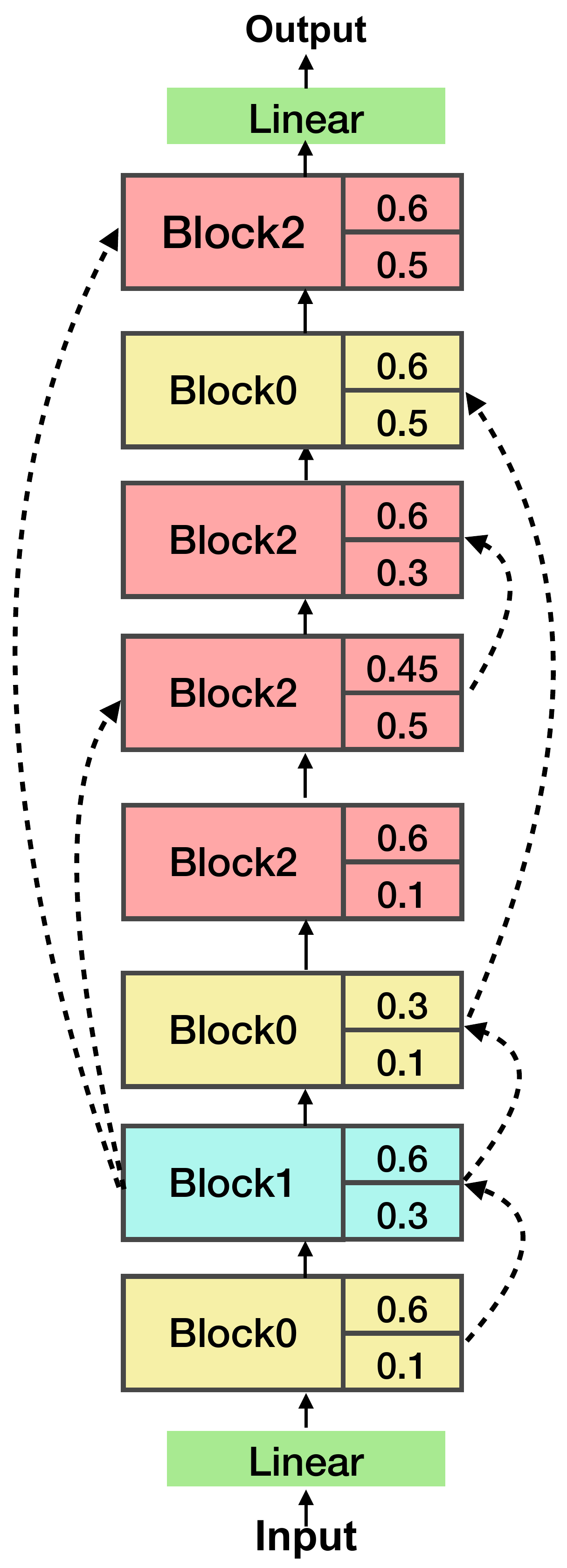}
\label{Model-CiteSeer}
}
\quad
\subfigure[Model-PubMed]{
\includegraphics[width=2.2cm]{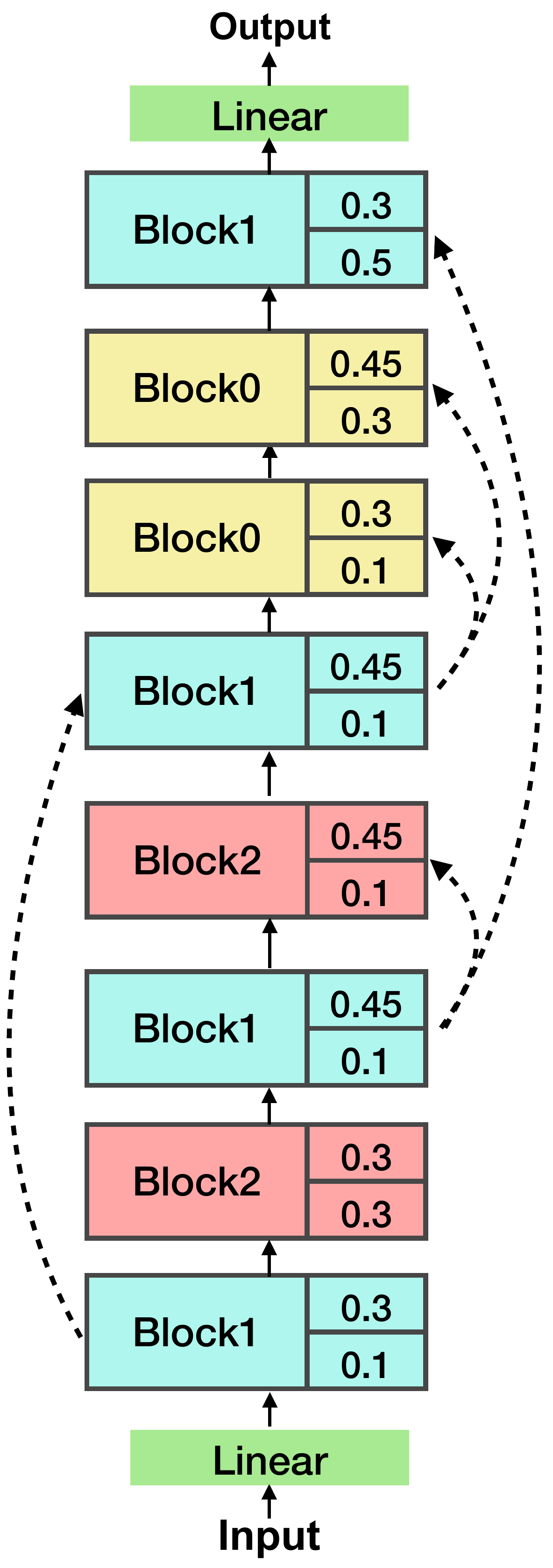}
\label{Model-PubMed}
}
\caption{(a). Our auto-generated top-3 GNN blocks from block-wise search towards semi-supervised task on Cora. (b), (c) and (d) are best architectures found in full-supervised tasks on Cora, CiteSeer and PubMed respectively.}
\end{figure*}

For the full-supervised learning tasks, we reuse GNN blocks obtained from semi-supervised settings on Cora and only execute the second-stage search on different datasets. We fix some of the hyper-parameters as follows: learning rate = $3\times10^{-2}$, weight decay for graph convolutional layers = $3\times10^{-3}$ and hidden dimension = 32. Besides three commonly used citation networks, we further introduce three web page datasets as in \cite{DBLP:conf/iclr/PeiWCLY20}: Cornell, Texas and Wisconsin. In these datasets, each node correspond to a web page and edges represent hyperlinks between pages. Node features are the bag-of-words representation of web pages. For fair comparsion, we split all the datasets following \cite{DBLP:conf/icml/ChenWHDL20}: 60\% nodes are set for training while 20\% for validation and the remaining 20\% for testing. We train each model for 400 epochs and report the results in Table \ref{full_table}. The generated architectures correspond to Cora, CiteSeer and Pubmed are depicted in Figure \ref{Model-Cora}, \ref{Model-CiteSeer} and \ref{Model-PubMed} respectively. The metrics for GCN, GAT and Genom-GCN are reused from \cite{DBLP:conf/iclr/PeiWCLY20}. From the results we could observe that our deep GNN model constructed by transferred blocks obtain the highest validation accuracy on all of the 6 datasets. Notably, our proposed method surpasses all other models in the last three datasets by large margin. (\ie over 14\% gain in each of the web page datasets). We assume such performance improvement come from not only our models depth, but also the connection and layer diversity and adaptivity, which is not considered by previous methods. We will further discuss and validate such assumption in section \ref{sec_diversity}.

\subsection{Model Transferability Test}
Even our transferred blocks show competitive performance on various datasets and tasks, we still want to further validate our models capability on pure transfer learning scenarios as proposed by \cite{DBLP:conf/ijcai/GaoYZ0H20}. Thus, in this section, we directly reuse the architecture obtained on citation networks and test it on supervised node classification tasks. Here we apply our model to four datasets: two coauthor datasets as MS-CS and MS-Physics, two product networks as Amazon Computers and Amazon Photo \cite{DBLP:journals/corr/abs-1811-05868}. Following \cite{DBLP:conf/ijcai/GaoYZ0H20}, 500 nodes in each dataset are selected for validation, 500 for testing while the remaining nodes are used for training.  We also reuse the metrics as reported in \cite{DBLP:conf/ijcai/GaoYZ0H20}. For parameter settings, we set the weight decay and hidden dimension for graph convolutional layers to $3\times10^{-2}$ and 32 respectively. Each model is trained for 800 epoch using Adam with learning rate 0.06. Classification results are shown in Table \ref{trans_table}. We run our model for 10 times and report the mean accuracies with variances. Consequently, our directly transferred GNN model reaches best performance on the four benchmarks, which further demonstrates that transferability not only lies in our generated blocks but also the final architectures.

%---------------------------------Discussion and Future Work---------------------------------

%---------------------------------Figure of Structural Diversity---------------------------------
\begin{figure}
\centering
\subfigure[CiteSeer]{
\includegraphics[width=3.9cm]{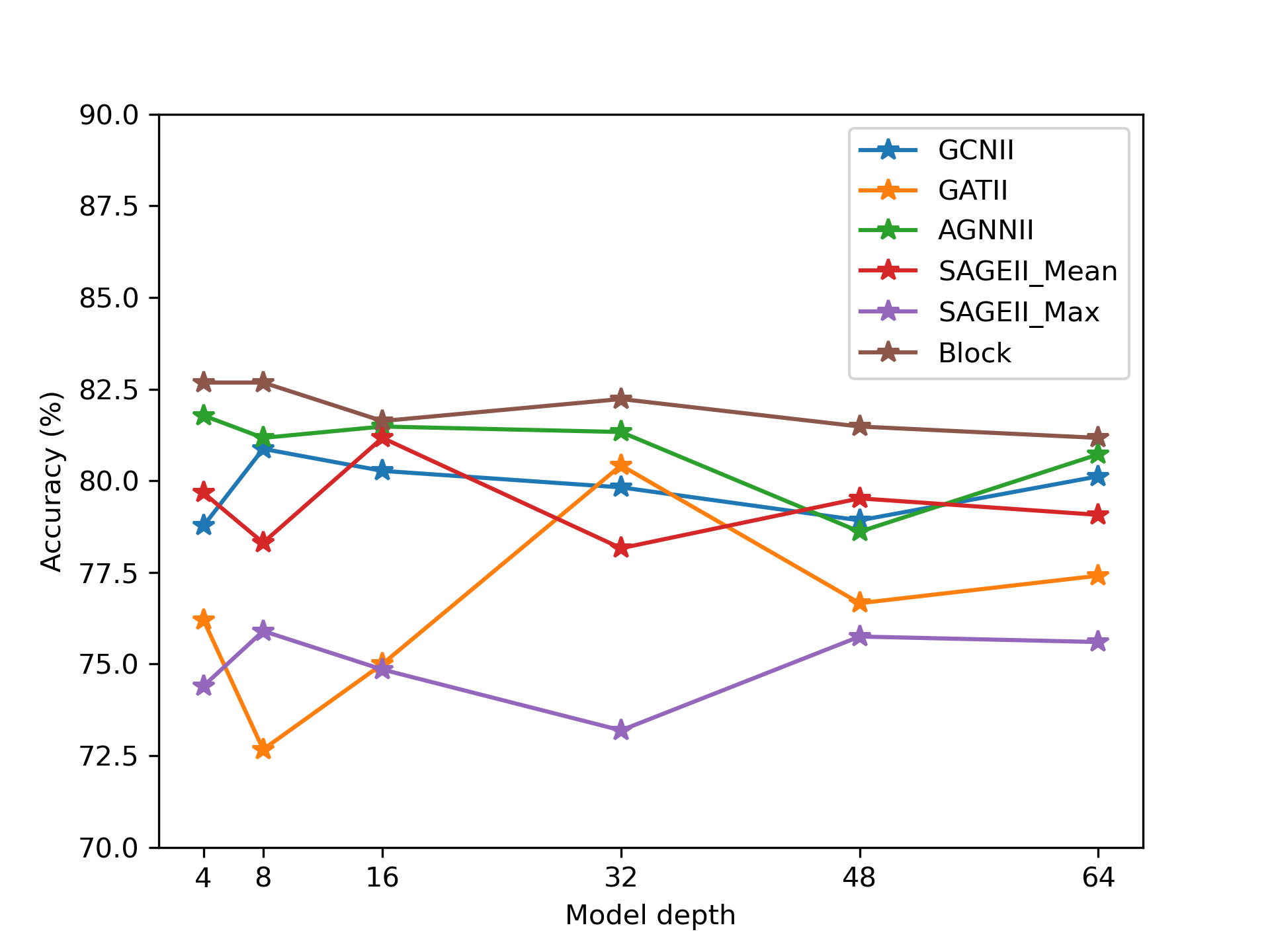}
\label{diversity-ablation-CiteSeer}
}
\subfigure[PubMed]{
\includegraphics[width=3.9cm]{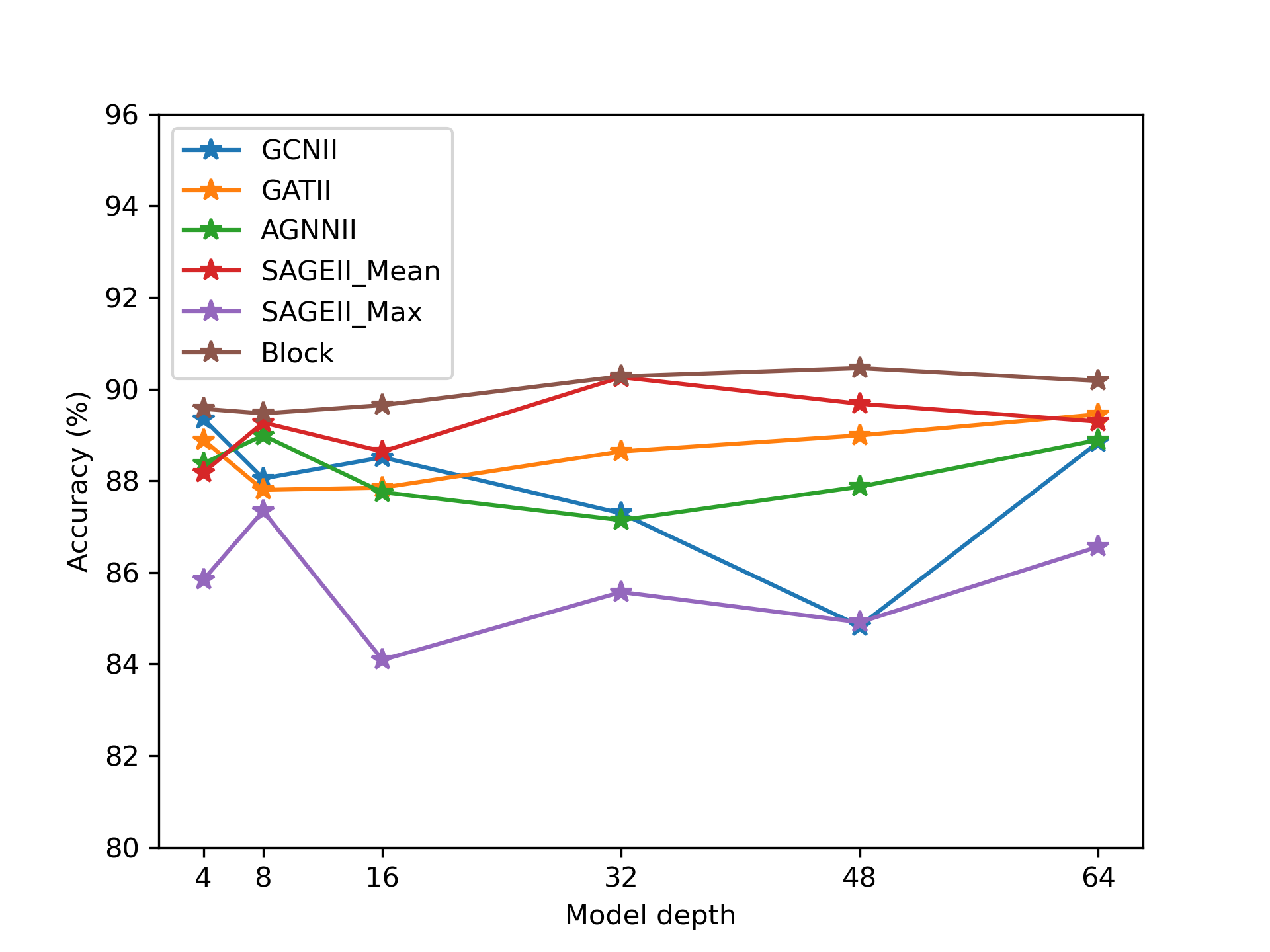}
\label{diversity-ablation--PubMed}
}
\caption{Ablation study on structural diversity with model depth increasing from 4 to 64.}
\label{Ablation}
\end{figure}

%---------------------------------Figure of DQN progressive---------------------------------
\begin{figure}
\centering
\subfigure[CiteSeer]{
\includegraphics[width=3.9cm]{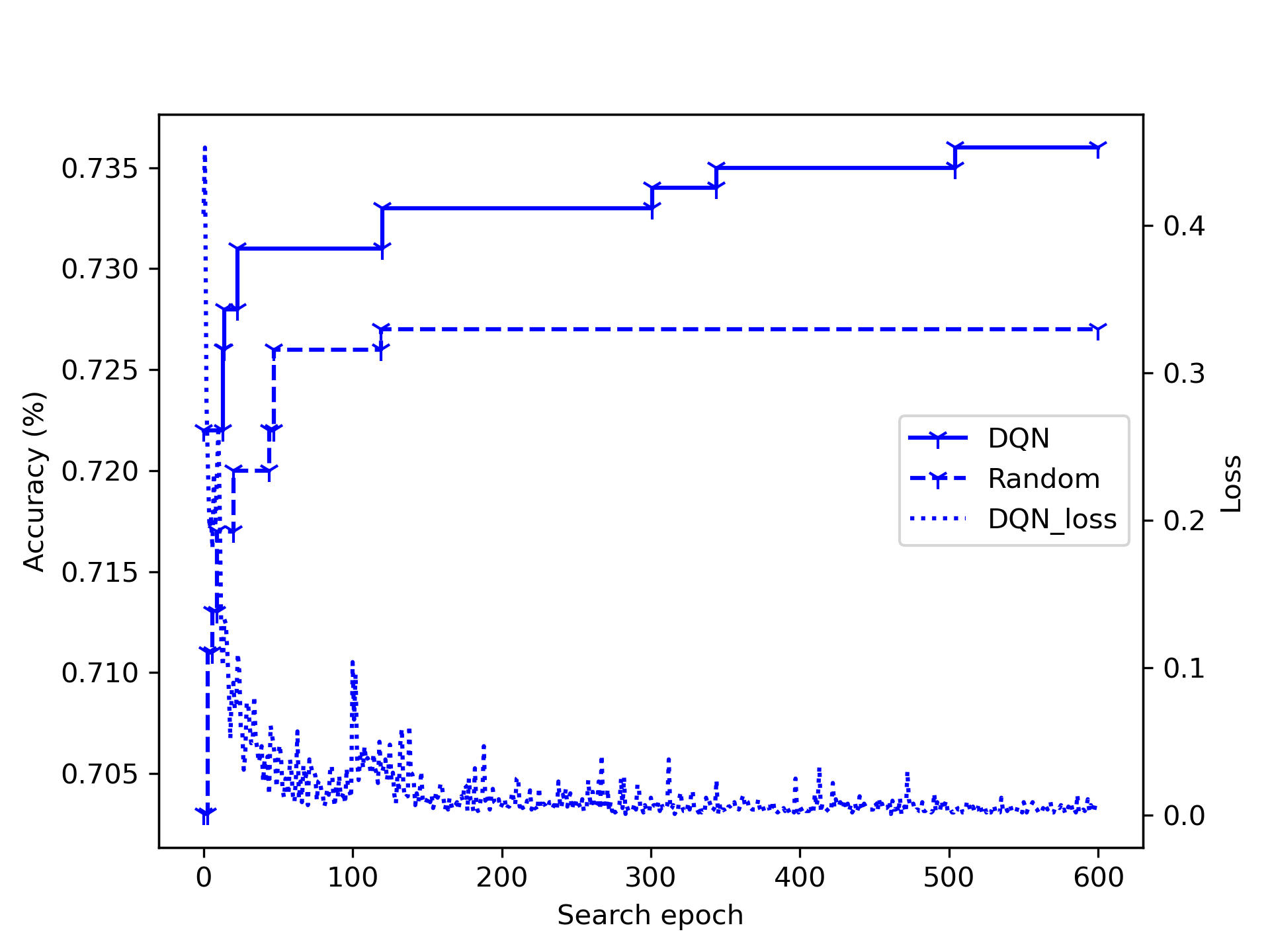}
}
\subfigure[PubMed]{
\includegraphics[width=3.9cm]{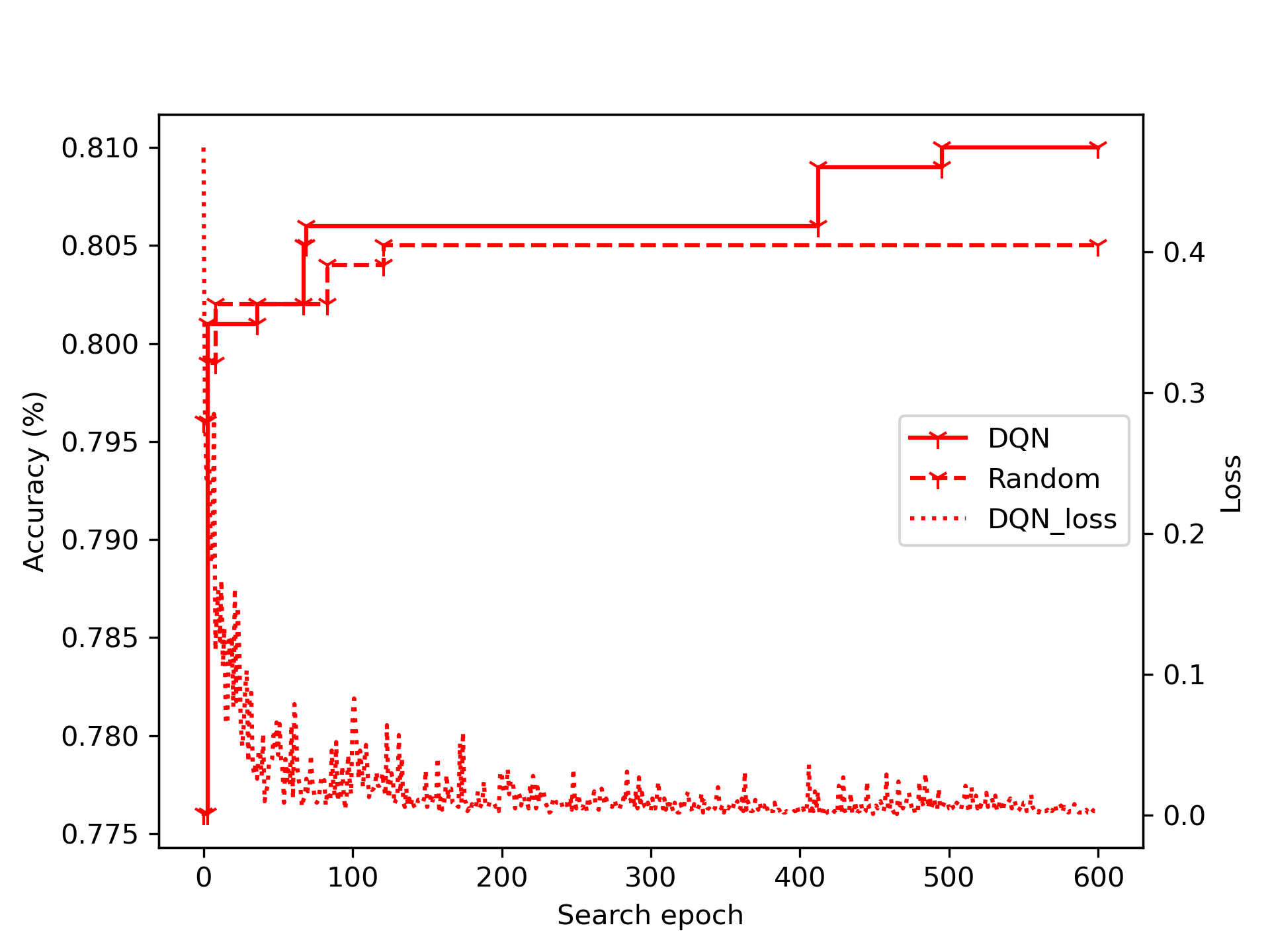}
}
\caption{Progression of obtained best model with DQN guided search and random search.}
\label{Progress}
\end{figure}

\subsection{Discussions and ablation studies}\label{sec_diversity}

\textbf{Importance of structural diversity.} Besides model depth, we assume our model's superiority comes from the layer diversity and flexibility of connections. To validate this, we conduct ablation study by fixing the layer type and connectivity of final architectures, and then comparing those fixed models with the standard architecture built upon our best block. For fair comparison, we keep hyper-parameters of all models consistent and report the mean accuracy of three independent trainings. As depicted in Figure \ref{Ablation}, our generated GNN block could retain stable and good performances when model depth changes, which demonstrates the benefits from structural flexibility and layer diversity.

\textbf{DQN iteration analysis.} Herein we study the effectiveness of our applied DQN reinforcement learning algorithm. We compare it with random search by visualizing the progression of obtained best architectures within a searching period of 600 epochs, alone with the loss of evaluation network in DQN, as depicted in Figure \ref{Progress}. We observe that, in the early stage (\ie explorations) our DQN agent behaves similarly to random search under the epsilon-greedy strategy. However, after the deep-q-network is sufficiently trained and the exploitation rate grows, our DQN agent could further find more high-performance architectures.

\section{Conclusion} 

In this paper, we study the problem of NAS-based deep graph neural network generations. We initially applied initial residual connection and identity mapping to various graph convolutional layers, and present a novel block-wise deep graph neural network generation pipeline. To efficiently explore the exponentially growing search space, we divide the overall search into two stages and utilize a deep-q-learning agent to guide the search process. In experiments, our generated deep GNN models reach state-of-the-art performance on various datasets and learning scenarios. Moreover, both our generated blocks and architectures show great transferability toward new datasets\footnote{Our code and trained model weight will be released in the final version for reproduction and comparison.}.

{\small
\bibliographystyle{ieee_fullname}
\bibliography{egbib}
}

\end{document}